\documentclass[letterpaper, 10 pt, conference]{ieeeconf}  

\IEEEoverridecommandlockouts                              

\usepackage{graphicx}
\usepackage{amsmath} 
\usepackage{amssymb}  
\usepackage{algorithmic, algorithm}
\usepackage{dsfont}

\newcommand{\numberset}[1]{\mathbb{#1}}
\newcommand{\Mat}{\numberset{M}}
\newcommand{\Lat}{\numberset{L}}
\newcommand{\Eat}{\numberset{E}}
\newcommand{\Rat}{\numberset{R}}
\newcommand{\Cat}{\numberset{C}}

\title{\LARGE \bf
HouseExpo: A Large-scale 2D Indoor Layout Dataset for Learning-based Algorithms on Mobile Robots
}

\author{Tingguang Li, Danny Ho, Chenming Li, Delong Zhu, Chaoqun Wang,
	 Max Q.-H. Meng*, \emph{Fellow}, \emph{IEEE} 
	\thanks{*The corresponding author.}
\thanks{Tingguang Li, Danny Ho, Chenming Li, Delong Zhu, and Chaoqun Wang are with the Department of Electronic Engineering, 
	The Chinese University of Hong Kong, Shatin, N.T., Hong Kong SAR, China. {\tt\small Email: \{tgli, dho, dlzhu, cmli, cqwang\}@ee.cuhk.edu.hk}. Max Q.-H. Meng is with the Department of Electronic and Electrical Engineering of the Southern University of Science and Technology in Shenzhen, China, on leave from the Department of Electronic Engineering, The Chinese University of Hong Kong, Hong Kong, and also with the Shenzhen Research Institute of the Chinese University of Hong Kong in Shenzhen, China. {\tt\small Email: max.meng@ieee.org}. }
\thanks{}%
}

\begin{document}

\maketitle
\thispagestyle{empty}
\pagestyle{empty}

\begin{abstract}
As one of the most promising areas, mobile robots draw much attention these years. Current work in this field is often evaluated in a few manually designed scenarios, due to the lack of a common experimental platform.
Meanwhile, with the recent development of deep learning techniques, some researchers attempt to apply learning-based methods to mobile robot tasks, which requires a substantial amount of data. 
To satisfy the underlying demand, in this paper we build HouseExpo, a large-scale indoor layout dataset containing $35,126$ 2D floor plans including $252,550$ rooms in total. Together we develop PseudoSLAM, a lightweight and efficient simulation platform to accelerate the data generation procedure, thereby speeding up the training process. 
In our experiments, we build models to tackle obstacle avoidance and autonomous exploration from a learning perspective in simulation as well as real-world experiments to verify the effectiveness of our simulator and dataset.
All the data and codes are available online and we hope HouseExpo and PseudoSLAM can feed the need for data and benefit the whole community.

\end{abstract}

\section{INTRODUCTION}
With the significant achievements in the AI field \cite{Silver2017Mastering}, the investigation of learning-based methods in the robotics area has received more and more attention in recent years.Many algorithms have been developed for mobile robots ranging from autonomous exploration \cite{Zhu2018Deep}\cite{8665177} to mapless navigation \cite{tai2017virtual}\cite{zhu2017target}.



From these achievements, the authors see the huge potential in applying learning-based methods to mobile navigation problems.
However, for the learning-based methods, the issue of data requirement must be addressed first.  The size and diversity of training data are crucial for the performance and will influence the generalization ability of the methods. Since the problems in robotics involve interactions with environments, getting data from real-world situations is impractical considering the resource and time cost. Therefore there is a need for a large-scale dataset and a high-performance simulator to speed up the training process. On the other hand, it is still challenging for current datasets and simulators to meet such demand. For the existing datasets of 2D environments, the size, as well as the variability, is limited \cite{bormann2016room}\cite{mielle2018method}\cite{aydemir2012}, which will adversely affect the algorithm's performance. As for the simulators, the processing time to build the map through Simultaneous Localization And Mapping (SLAM) is time-costing, which is a bottleneck in training the neuron networks which routinely involve millions of trial-and-error episodes. These issues motivate the authors to develop a large-scale dataset, HouseExpo, and a fast simulation platform, PseudoSLAM, to improve the training efficiency.



\begin{figure}
	\centering
	\includegraphics[width=250pt]{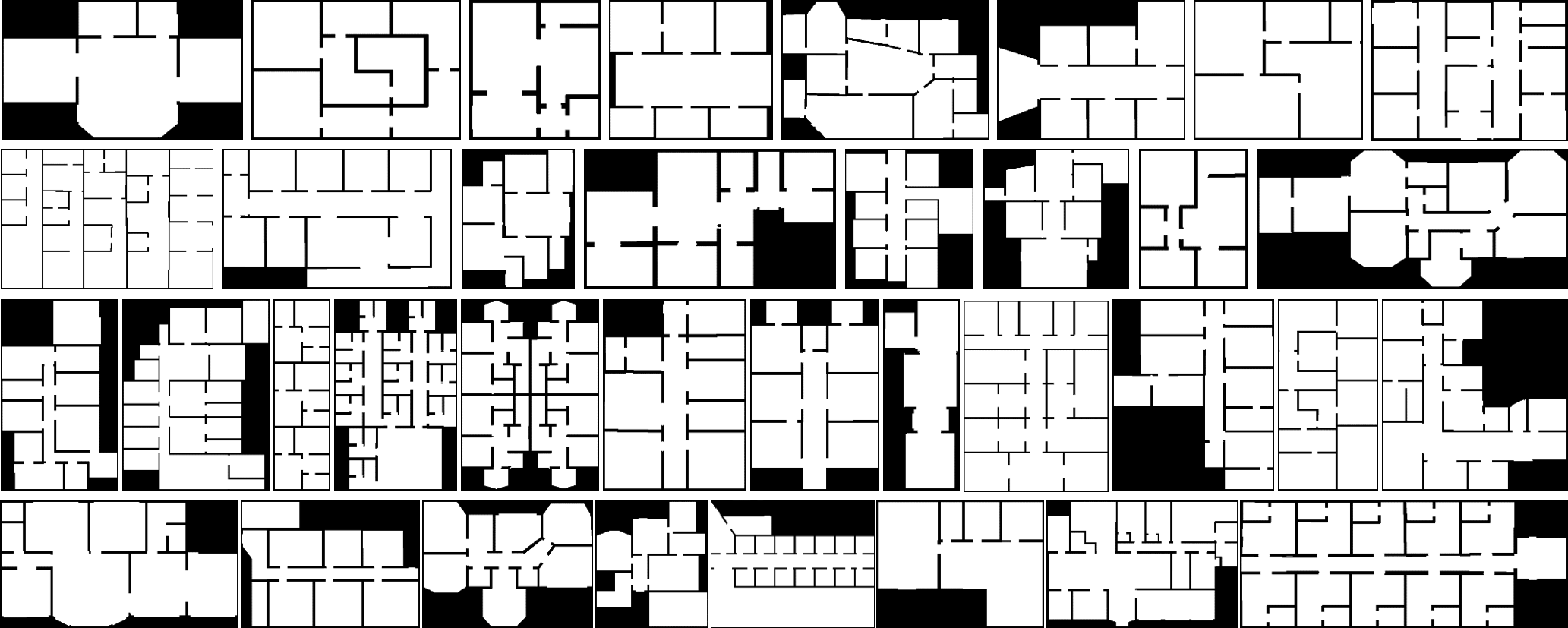}
	\caption{Some house samples from HouseExpo dataset. Black pixels mean obstacles and white pixels represent free space.} 	\label{fig:dataset}
\end{figure}

HouseExpo\footnote{The HouseExpo dataset and the simulation platform code are available at {\tt\small https://github.com/TeaganLi/HouseExpo/}} is a large-scale 2D floor plan dataset built on SUNCG dataset \cite{song2016ssc}, consisting of $35,126$ human-designed 2D house blueprints including $252,550$ rooms in total, ranging from single-room studios to multi-room houses (some map samples are displayed in Fig. \ref{fig:dataset}). The details of the dataset generation pipeline are presented in Section \ref{sec:dataset}. 

PseudoSLAM is a lightweight simulation platform with OpenAI Gym-compatible interface \cite{brockman2016openai} which simulates SLAM and the navigation process in an unknown 2D environment. It reads the data from HouseExpo, creates the corresponding 2D environment and generates the mobile robot to carry on different tasks in this environment. The detailed introduction is given in Section \ref{sec:simulator}. 

To demonstrate the effectiveness and the efficiency of HouseExpo and PsdudoSLAM, we re-examine two tasks: object avoidance and autonomous exploration, based on Deep Reinforcement Learning (DRL) in the experiment part. We also implement a real-world experiment on a \emph{TurtleBot}. The result shows that the knowledge learned in PseudoSLAM can be transferred to the real world without additional fine-tuning and the indoor spatial structure can help guide the exploration process.


In summary, our work has the following contributions:
\begin{itemize}
	\item A large-scale 2D indoor layout dataset with a diverse spatial structure is built, containing $35,126$ maps.
	\item A high-speed simulation platform is developed to improve the training efficiency of DRL network.
	\item The effectiveness of HouseExpo and PseudoSLAM is verified via simulation and real-world experiments.
\end{itemize}

\section{HouseExpo DATASET} \label{sec:dataset}
In recent years, many researchers attempt to apply deep learning techniques to mobile robots. However, one of the challenges of training deep neural networks is the lack of large datasets with diverse samples. On the one hand, the sizes of the existing 2D floor plan datasets are limited. As far as we know, the largest 2D floor plan datasets are MIT campus dataset \cite{whiting2007} and KTH campus dataset \cite{aydemir2012} with the size of $775$ and $165$ floor plans, respectively. Apart from their limited size, the lack of diversity of their samples is another concern. Both MIT and KTH datasets are collected from campus buildings and thus the location of rooms obeys some particular pattern, e.g. rooms in campus buildings are often orderly arranged along corridors. This pattern may not apply to other scenarios like household environments, and limits their variety of samples. Besides, neither of these two datasets considers the importance of the connectivity between rooms. Many rooms are isolated and a robot may be initialized at these isolated rooms and trapped there rather than exploring the whole environment. Furthermore, neither of them provides easy-to-use tools to load the floor plans into simulators, making it difficult to use them for other applications. 


Due to the lack of a simple yet diversified floor plan dataset, current work is evaluated either in simple simulated environments \cite{bai2017toward}\cite{wang2017towards}, lacking realism in terms of spatial structure, or in a limited number of similar scenes \cite{zhu2017target}, deficient in verifying generalization capacity. In view of this, we create HouseExpo dataset consisting of $35,126$ environments with a total of $252,550$ rooms to benefit the investigation of data-driven approaches. 

\subsection{Dataset Generation}
\begin{figure}
	\centering
	\includegraphics[width=200pt]{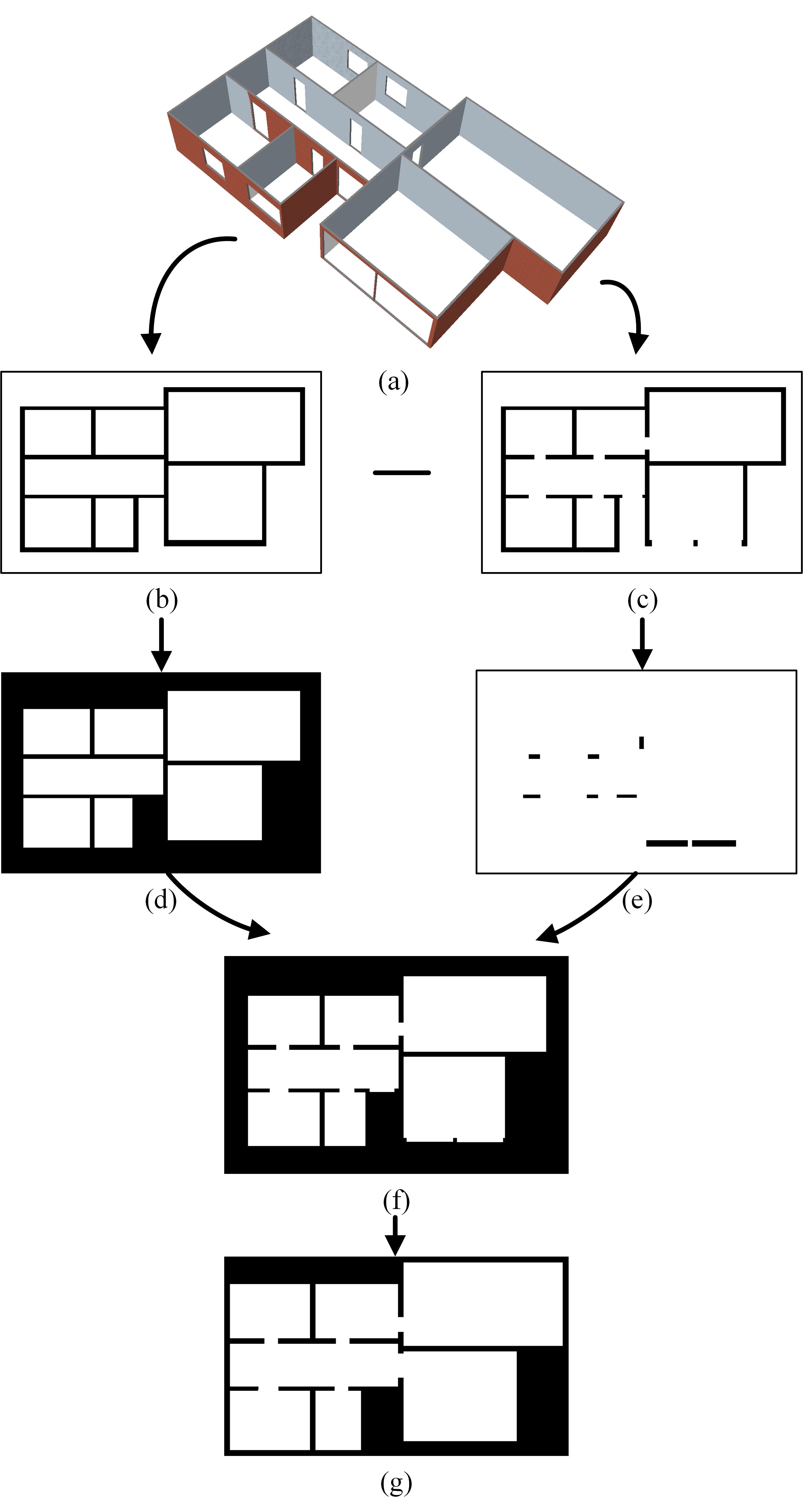}
	\caption{The pipeline of generating HouseExpo. (a) A 3D model $s_i$ from SUNCG dataset. (b) The ground cross section $\mathcal{P}_g$. (c) The door cross section $\mathcal{P}_d$. (d) Find the boundary of $\mathcal{P}_g$ and fill the outside as obstacles (denoted as $\mathcal{P}_{g'}$). (e) The location information of doors $\Lat$ is obtained via subtracting $\mathcal{P}_d$ from $\mathcal{P}_g$. (f) Remove the doors in $\mathcal{P}_{g'}$ based on $\Lat$. (g) Final result after cell-filling, connectivity-checking, line refinement and  image cropping.}  	\label{fig:pipeline}
\end{figure}

\begin{algorithm}[h]
	\caption{2D Floor Plan Dataset Generation} \label{alg:generation}
	\begin{algorithmic}[1]
		\renewcommand{\algorithmicrequire}{\textbf{Input:}}
		\renewcommand{\algorithmicensure}{\textbf{Output:}}
		\REQUIRE Number of SUNCG models $k$,  ground plane height $h_{g}$, door plane height $h_{d}$, number of cell-checking samples $n$, number of connectivity-checking samples $m$, area threshold $\delta_a$;
		\ENSURE The set of 2D floor plans $\Mat$;
		\STATE $\Mat=\{\}, i = 0, j=0$;
		\FOR {$i=1$ to $k$}
		\STATE extract a 3D model $s_i$ from SUNCG;
		\STATE $\mathcal{P}_g = s_i(z=h_g)$, $\mathcal{P}_d = s_i(z=h_d)$;
		\STATE $\Lat = \{\boldsymbol{u}\in \Rat^2: \mathcal{P}_g(\boldsymbol{u}) \neq \mathcal{P}_d(\boldsymbol{u})\}$, $\boldsymbol{u}$ is ($x,y$) coordinate;
		\STATE calculate the contour of $\mathcal{P}_g$, fill in its outside as obstacles and get $\mathcal{P}_{g'}$;
		\STATE $\mathcal{P}_{g'}(\boldsymbol{u})=$ free, for all $\boldsymbol{u}$ in $\Lat$;
		
		\STATE calculate all the contours $\Cat$ of $\mathcal{P}_{g'}$;
		\STATE fill in $C$ if area($C$)$<\delta_a$, for all $C$ in $\Cat$;
		
		\STATE $\{\boldsymbol{u_1},...,\boldsymbol{u_m}\}=$UniformRandom($\Eat$, $m$), where $\Eat=\{\boldsymbol{u}\in \Rat^2: \mathcal{P}_{g'}(\boldsymbol{u})= \text{free}\}$;
		\STATE $\boldsymbol{M_{ij}}$=dist($\boldsymbol{u_i}, \boldsymbol{u_j}$) for all $i,j \in \{0...m\}$;
		\REPEAT
		\STATE pick $\boldsymbol{u_i},\boldsymbol{u_j}$ that $\boldsymbol{M_{ij}}=min(\boldsymbol{M})$;
		\REPEAT 
		\STATE plan a path $\xi$ between $\boldsymbol{u_i}$ and $\boldsymbol{u_j}$;
		\IF {$\xi = \varnothing$}
		\STATE remove a cross segment between the wall and the line($\boldsymbol{u_i}$,$\boldsymbol{u_j}$);
		\ENDIF
		\UNTIL {$\xi \neq \varnothing$}
		\UNTIL {all point pairs are checked}
		\STATE refine the wall and crop the generated map $\mathcal{P}_{g'}$;
		\STATE $\Mat= \Mat \cup \{\mathcal{P}_{g'}\}$;
		\ENDFOR
		\STATE compute the similarity of $\Mat$ and remove duplicate elements;
		\RETURN $\Mat$
	\end{algorithmic}
	\label{lidarPoseEstimate}
\end{algorithm}

The computer vision community witnessed a prosperity in 3D indoor datasets, e.g. SUNCG dataset~\cite{song2016ssc}, Matterport3D~\cite{Matterport3D} and Gibson~\cite{xiazamirhe2018gibsonenv} dataset, that are mainly designed for vision-related tasks. We build our HouseExpo dataset on SUNCG. The SUNCG dataset, consisting of $45,622$ manually-designed 3D house models, is originally created to facilitate semantic scene completion, a task for simultaneously producing a 3D voxel representation and semantic labels by using a single-view observation. The various human-designed house models in SUNCG make it suitable as a base for our application.

However, there are several issues if we directly utilize the SUNCG dataset. First of all, similar to KTH and MIT dataset, the SUNCG dataset does not guarantee the connectivity among rooms.
In addition, some mobile robot tasks can be treated as 2D problems \cite{Zhu2018Deep}\cite{tai2017virtual}\cite{chiang2015path}. Using 3D environments in SUNCG for 2D task algorithms is inefficient or even inapplicable.
Furthermore, as the SUNCG dataset is originally designed for scene completion task, it involves too much semantic information like textures, which incurs additional computational costs.   

To satisfy the demand for a large-scale 2D layout dataset, we built HouseExpo dataset. An illustrative example of our pipeline to generate the dataset is shown in Fig. \ref{fig:pipeline}. First, we extract a 3D structure model $s_i$ from SUNCG dataset. Note that the projection of the top view of $s_i$ onto a 2D plane cannot be regarded as the desired indoor map since it fails in reflecting the connectivity relationship between rooms due to the existence of the lintel. Then we obtain the ground cross-section plane $\mathcal{P}_g$ and the door cross-section plane $\mathcal{P}_d$ at the height of $h_g$ and $h_d$, respectively. As a result, the door location set $\Lat$ can be easily determined by subtracting $\mathcal{P}_d$ from $\mathcal{P}_g$. In addition, considering that the rooms in $\mathcal{P}_g$ are closed we can readily calculate the contour of the house according to \cite{suzuki1985topological} and get the indoor layout $\mathcal{P}_{g'}$ by filling in the outside of the boundary as obstacles. Further, the doors are removed from $\mathcal{P}_{g'}$ with the knowledge of door location set $\Lat$. 

Notice that the cross section of 3D models may contain some closed cells, caused by cross sectioning particular regions like chimneys and unused space between rooms, which are meaningless and must be removed. To remove such cells, we calculate all the room contours using \cite{suzuki1985topological} and then fill in the small cells as obstacles. Furthermore, to tackle the connectivity issue, we recheck the connectivity in every generated map: (1) first we uniformly sample $m$ points on $\mathcal{P}_{g'}$ and compute the 
Euclidean distance of each point pairs, denoted by the distance matrix $\boldsymbol{M}$; (2) then we pick the two points $\boldsymbol{u}$ and $\boldsymbol{v}$ with the shortest distance in $\boldsymbol{M}$ and plan a path between them using A* algorithm \cite{hart1968formal}; (3) if step (2) fails, indicating the two areas are not connected yet and manually establishing the connectivity is necessary, we will remove a wall segment which intersects with the line between $\boldsymbol{u}$ and $\boldsymbol{v}$; (4) repeat step (2) until all point pairs are checked. 

One remaining step is to refine the wall appearance to maintain the consistent thickness along the same wall and crop the image to center the house. Incidentally, there exist some duplicate samples in our generated dataset due to the characteristic of SUNCG, where each scene is a combination of house models and various objects. To reduce such redundancy, we re-examine the similarity among maps by simply calculating their pixel difference and remove the duplicate ones. The detailed introduction is given in Algorithm \ref{alg:generation}, where $k=45622$, $h_g=0.01$m, $h_d=0.2$m, $n=500$, $m=100$ and $\delta_a=2\text{m}^2$. 

The above floor plan generation process is dataset-agnostic and can be applied to any 3D house models. While due to the diversity of SUNCG, human labeling is involved for some extreme cases: (1) for some isolated rooms far from main areas like garages, they are regarded as inaccessible and erased from the map; (2) for some scenes where there are no layout but only objects, we exclude them from our HouseExpo dataset; (3) for some open/semi-open houses where there is no wall as their boundaries, we design walls for them to make sure every house has a distinct boundary; (4) for some houses with unreasonable layout, for example, the walls are shaped as some characters, we remove them from our dataset. 

\subsection{The Statistics Information} 
\begin{figure}
	\centering
	\includegraphics[width=0.50\textwidth]{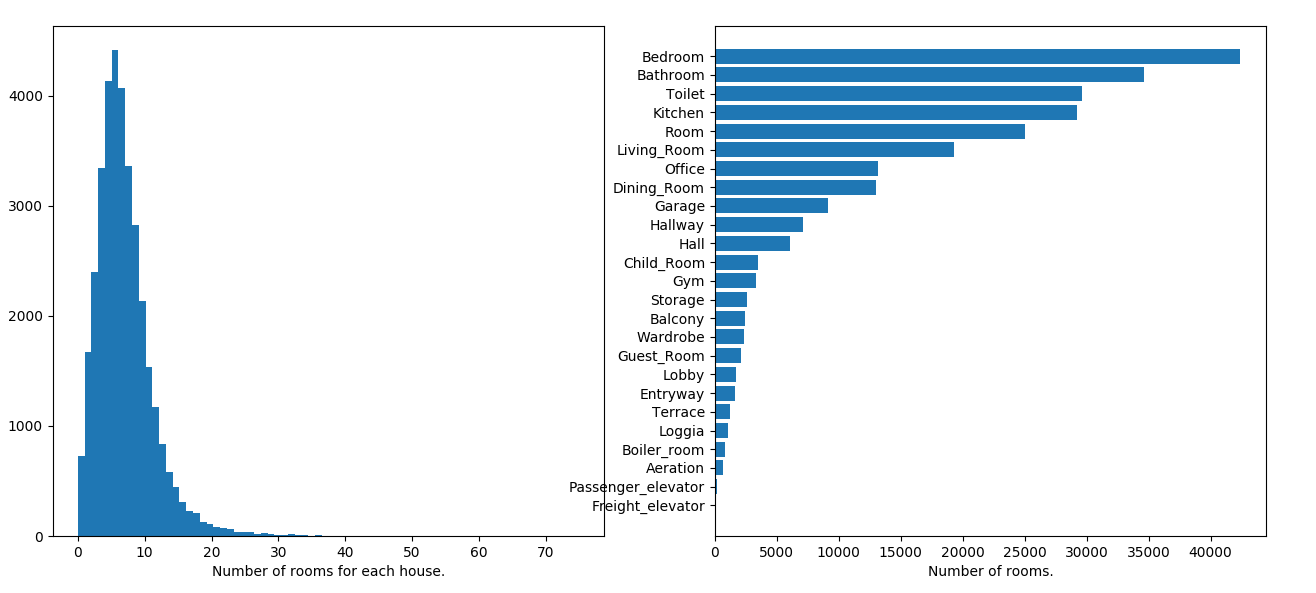}
	\caption{Room number distribution and room category label distribution.} 
	\label{fig:distribution}
\end{figure}

Instead of storing floors plans as images, we store them as JSON files where the structure information is represented as line segments with respect to the house's centroid coordinates in meter, aiming to make HouseExpo more realistic. There are $35,126$ houses with $252,550$ rooms, with a mean of 7.14 and a median of 7.0 rooms per house. Furthermore, the room category labels are inherited from SUNCG dataset, aiming to provide semantic information. The distribution of rooms per house and room category labels are displayed in Fig. \ref{fig:distribution}.   

\section{PseudoSLAM SIMULATION PLATFORM} \label{sec:simulator}

\subsection{Motivation}
For learning-based algorithms on mobile robots, the typical observations are sensory data, e.g. RGB image~\cite{zhu2017target} or LiDAR scan~\cite{tai2017virtual}. However, due to the gap between simulation and reality, models trained in simulators often require fine-tuning in the real world. This inspires us to utilize occupancy grid map as observation. The occupancy grid map is an output form of the SLAM process and it is uniform in both simulation and reality. Therefore, the discrepancy problem between simulators and the real-world can be avoided.


SLAM algorithms build occupancy grid maps. However, the computational complexity of SLAM algorithms makes them extremely inefficient for learning-based methods which usually require a substantial amount of data at the training stage. In light of this, we develop PseudoSLAM to simulate the SLAM process. Instead of using sensory data, it directly utilizes the ground truth map to generate the map. Consequently, it can achieve a competitive mapping result as SLAM but at a much lower time cost. 


\subsection{Simulator Overview} \label{sec:sim_overview}
PseudoSLAM abstracts away low-level sensor data processing so that users can focus on performing high-level strategic policy based on the built map.
When developing the simulation platform, we have two objectives: 
(1) Time efficiency and low computational cost; 
(2) Close to real-world situations such that the model developed using the simulator can be transferred to the real world.
Apart from simulating the SLAM process, the simulation platform is capable to generate obstacles, aiming to increase the variance of training samples and narrow its gap with the real-life scenarios. Besides, PseudoSLAM provides an OpenAI-Gym compatible interface so that users can easily integrate existing learning-based methods. Furthermore, our simulator provides some parameters that users can specify (as listed in Table~\ref{tab:parameter}), aiming to satisfy the need for a wider range of applications.



\begin{table}[]
     \centering
     \caption{Some Parameters in PseudoSLAM that Users can Specify}
        \begin{tabular}{l|l}
            \hline \hline 
            Parameters  &  \\
            \hline
            Robot configuration & Robot size, step length (linear and angular)\\
            Mode & Exploration in unknown environments or \\
                 & navigation in fully known environments \\
            Obstacle & User defined or randomly generated \\
            Resolution & Ration between meter and pixel \\
            Sensor & Range, field of view, noise \\
            Noise & Linear and angular registration error\\
            \hline
        \end{tabular}
     \label{tab:parameter}
 \end{table}
 
\subsection{PseudoSLAM Pipeline} \label{sec:pipeline}
The PseudoSLAM aims to simulate SLAM algorithms with the knowledge of the ground truth of the map.
Its map format is consistent with that of the SLAM algorithms in ROS gMapping~\cite{grisetti2007improved} and hector slam package \cite{kohlbrecher2011flexible}. It is an occupancy grid map, consisting of three states, i.e. free space, obstacle and uncertain areas, represented by different pixel values.

\begin{figure}
	\centering
	\includegraphics[width=250pt]{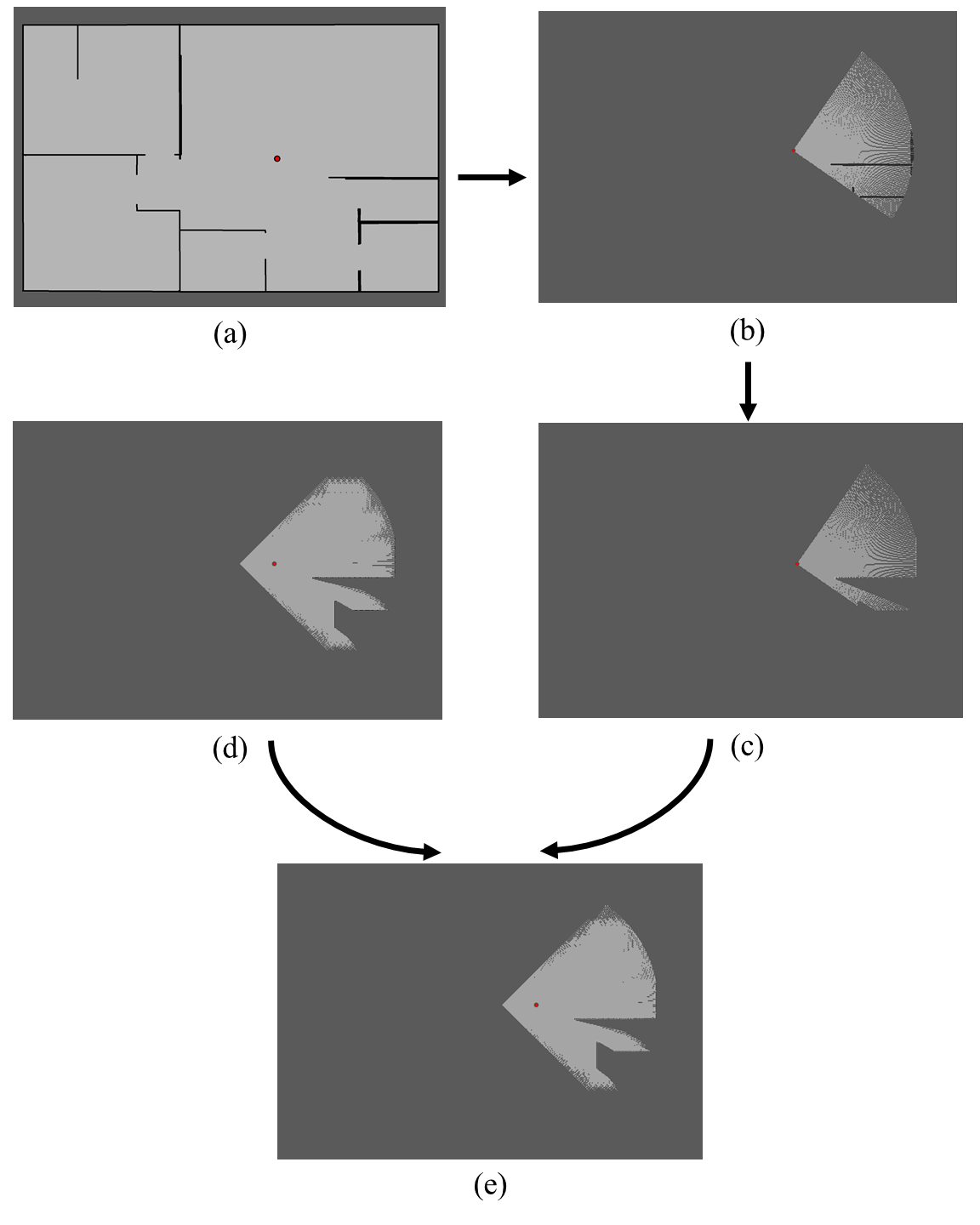}
	\caption{The pipeline of PseudoSLAM. (a) The ground truth map $\mathcal{G}$. (b) The cropped sector $\mathcal{S}_t$ centered at the robot $x_t$ at time $t$. (c) The processed sector $\mathcal{S'}_t$. (d) The occupancy map $\mathcal{M}_{t-1}$ at time $t-1$. (e) The occupancy map $\mathcal{M}_t$ at time $t$.}
	\label{fig:PseudoSlam_pipeline}
\end{figure}

The workflow of PseudoSLAM is shown in Fig. \ref{fig:PseudoSlam_pipeline}:
(1) A floor plan is loaded as the ground truth map $\mathcal{G}$;
(2) The robot pose $x_t$ is updated according to its motion, and a sector $\mathcal{S}_t$ centered at $x_t$ with a radius of the laser range and angle of the field of view is cropped; 
(3) Process $\mathcal{S}_t$ to hide the areas behind obstacles. The obstacle locations in the sector are identified first. Along the robot-obstacle line, pixels behind the obstacle are set uncertain, and pixels between the robot and obstacle are set free. The processed sector is denoted as $\mathcal{S'}_t$; 
(4) Merge $\mathcal{S'}_t$ into the occupancy map $\mathcal{M}_{t-1}$ built at $t-1$ and obtain $\mathcal{M}_t$.
The whole process is repeated when the robot is exploring an environment.



\subsection{Noise and Uncertainty Model} \label{sec:noise}
Using the above pipeline, an ideal SLAM result will be generated. However, real environments are full of noise and uncertainty. Therefore, we explicitly add noise and uncertainty to the simulator to minimize such gap. Here, we simulate the noise of laser range measurement and the laser point matching and registration uncertainty \cite{schaefer2017analytical}\cite{petrovskaya2009model}. Besides, we assume that the total reflection of the laser pulse does not exist.

\subsubsection{Laser Scan Noise}
There always exists noise in the measured phase, and the laser noise is proposed to be modeled as the Gaussian distribution in the literature \cite{schaefer2017analytical}\cite{petrovskaya2009model}. To simulate the noise, each obstacle point is shifted by $x$ pixel along the robot-obstacle segment, where $x$ is sampled from the Gaussian distribution with a mean of 0 and a user-defined standard deviation.

\subsubsection{Matching \& Registration Uncertainty}
As for the SLAM, there may be matching error when registering laser points to the global map, causing a shift of the observation sector in the map. 
To simulate this phenomenon, the processed sector is rotated by $\theta$ and transformed by a linear shift of $(x,y)$ in each step, while $\theta$ and $(x,y)$ is sampled from the Gaussian distribution with a mean of $0$ and a user-defined standard deviation.


\subsection{Obstacle Generation} \label{sec:obstacle}
As mentioned in Section \ref{sec:sim_overview}, adding obstacles in the simulator can help to increase the variance of training data, i.e. improving the diversity of floor plans. 
On the other hand, most houses in the real world are filled with furniture, and adding obstacles in our simulator can help make the model more adaptable to real world environments.


The obstacles can be either specified by users or randomly generated by the simulator. Users can explicitly specify the number of obstacles, obstacle shapes and locations. The obstacles can also be generated at random locations by the simulator, without overlapping with themselves or with the walls.
Apart from static objects, users can also add dynamic obstacles by providing their trajectories. This is useful for simulating environments populated with moving people.

\subsection{Comparison} \label{sec:comparison}
\subsubsection{Time efficiency}
\begin{figure}
	\centering
	\includegraphics[width=200pt]{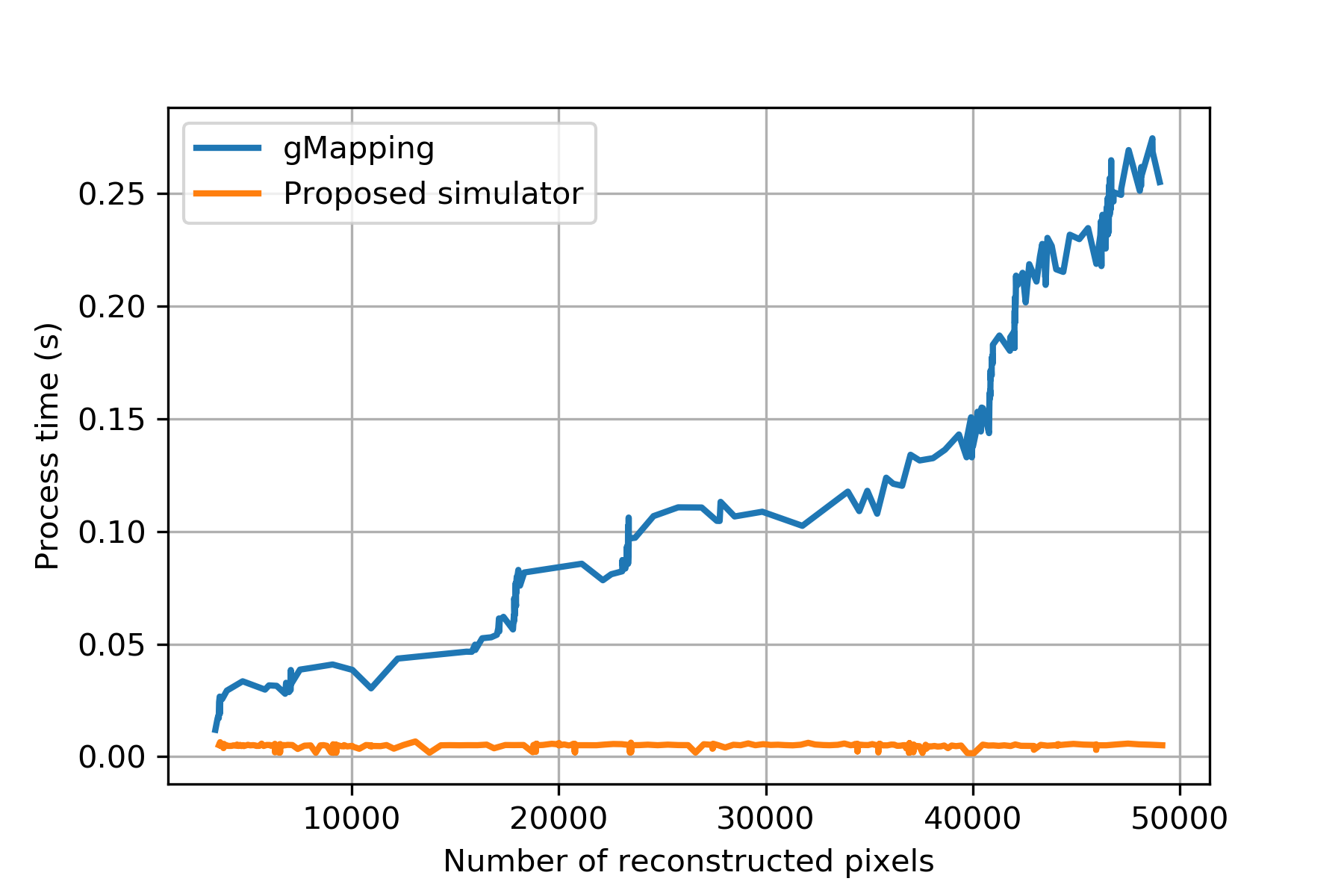}
	\caption{Process Time Comparison. The PseudoSLAM simulator and gMapping are tested on a laptop with Intel Core i5-6500.}  	 
	\label{fig:sim_speed}
\end{figure}

We test the time efficiency by comparing the map building processing time using gMapping~\cite{grisetti2007improved}, and PseudoSLAM (as shown in Fig.~\ref{fig:sim_speed}). It can be seen that the process time of the developed simulator is much shorter and the increment of map size does not affect the mapping speed. 

\subsubsection{Map Similarity}
\begin{figure}
	\centering
    \includegraphics[height=0.28\textwidth, width=0.45\textwidth]{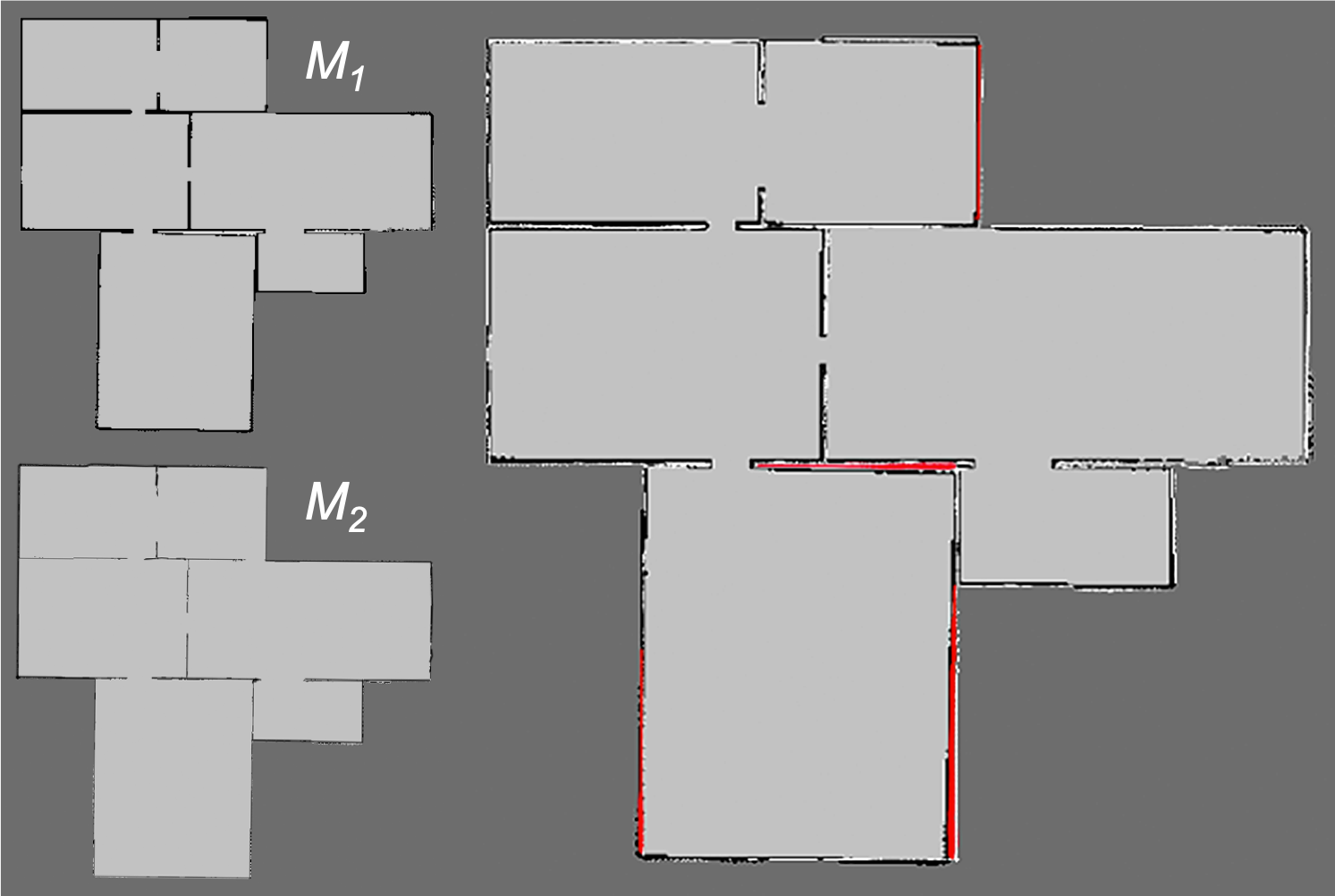}
	\caption{Map similarity comparison. Left top map, denoted as $M_1$ is built by gMapping in Stage and left bottom map denoted as $M_2$ is the output by our simulator. Right map is the overlapped result where walls from $M_1$ are in black color and walls from $M_2$ are in white color. The mismatched areas are marked in red color.} 
	\label{fig:sim_slam}
\end{figure}
To test how close is our simulation result to the map generated by SLAM algorithms, we evaluate their similarity.  One floor plan is picked from HouseExpo and used to build the occupancy map using gMapping~\cite{grisetti2007improved} in Stage~\cite{vaughan2008massively} and PseudoSLAM, respectively, as shown in Fig.~\ref{fig:sim_slam}. The left top map is the result of gMapping and the left bottom map is the result of PseudoSLAM. The right map is the overlapped result where mismatched areas are highlighted in red color. To compare their similarity, we compute their Intersection over Union (IoU) where the walls are regarded as the bounding box in terms of indoor spaces. Our simulator achieves $98.2\%$ of IoU, indicating the result of our simulator is competitive to the output of SLAM algorithms used in the real world.

 \section{EXPERIMENTS} \label{sec:experiment}
 \subsection{Experiment Setup}
 
 In this section, we report experimental results regarding the following questions: (i) How well does the policy trained in PseudoSLAM generalize to real-world situations? And (ii) whether the spatial structure information in HouseExpo can be helpful in accelerating the exploration process?
 
 We implement two tasks, i.e. obstacle avoidance and autonomous exploration, to answer above questions. For the obstacle avoidance, we train a model in our simulator, then transfer the learned policy to a \emph{TurtleBot} robot platform without additional fine-tuning. The robot can navigate without collision in the room full of obstacles, showing the knowledge acquired from PseudoSLAM can be generalized to reality with zero refinement. For the autonomous exploration, we train DRL policies with floor plans from HouseExpo and test them in new environments. The result shows the space information can speed up the exploration process. 
 
 
 Both problems are formulated as Markov decision-making process, where at time $t$, the agent observes a state $s_t$, based on which it makes an action $a_t$, and receives a reward $r_t$ accordingly. In our model, $s_t$ is $l\times l\text{ m}^2$ rectangular area centered at the robot's position and its orientation is the same as robot's orientation, and $\{a_t\}$ corresponds to three directional movements \{\emph{forward}, \emph{left rotation}, \emph{right rotation}\}. 
 The agent is equipped with a laser with a range of $l_l$ meters and a horizontal field of view $l_a$ degrees.
 
 \subsection{Obstacle Avoidance} \label{sec:OA}
 In this part, we train a model in PseudoSLAM aiming to navigate the robot without collision through obstacle-filled environments and test the learned policy on a \emph{TurtleBot} in the real world. The real-world experiments are conducted in five scenes and results show that the knowledge learned in our simulator can be directly transferred to real robots without any fine-tuning, verifying the capability of our simulator.
 
 In our experiment, the goal of obstacle avoidance is to prevent the agent from hitting objects or walls, meanwhile cover as a long distance as possible. Therefore, the reward function $r_t$ is defined as 
 \[
 r_t = \begin{cases}
 -1.0, & \text{if a collision happens at $t$,} \\
 \alpha_s r_s + \alpha_a r_a, & \text{if no collision happens at $t$.} \\
 \end{cases}
 \]
If collision happens at time $t$, the agent receives a penalty of $-1.0$. Otherwise, the reward is the weighted sum of state reward $r_s$ and action reward $r_a$. The state reward $r_s$ is the newly discovered area at time $t$, encouraging the agent to move towards unknown areas. The action reward is defined as $r_a=1_{forward}(a)$, where $1$ is an indicator function with a value of $1$ if $a_t$ is \emph{forward} and $0$ otherwise, preventing the cases where the agent keeps rotating in place. In our case, $\alpha_s=0.9$ and $\alpha_a=0.1$.
 
\begin{figure}
	\centering
	\includegraphics[width=0.5\textwidth]{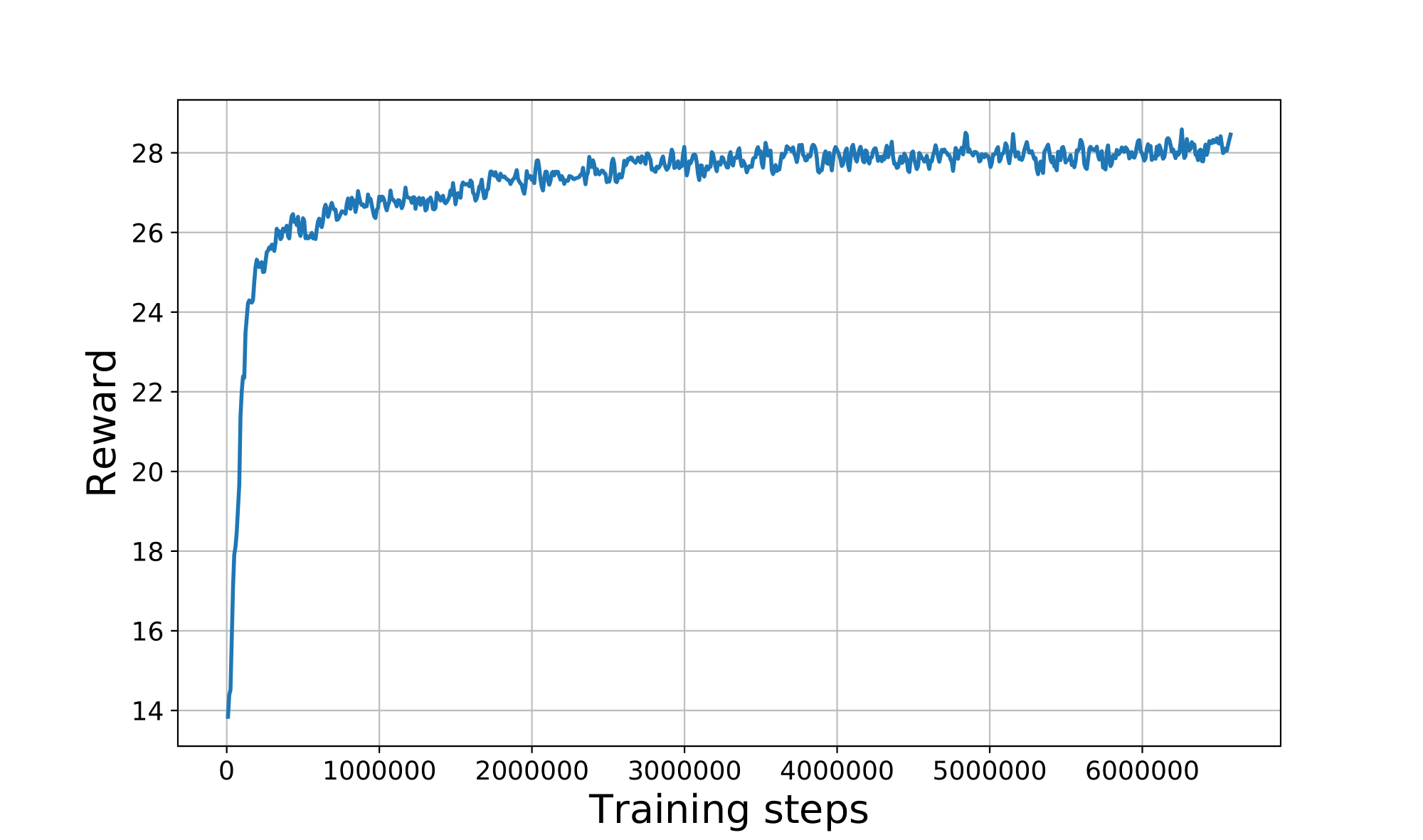}
	\caption{The learning curve of learning-based obstacle avoidance at training stage. } 
	\label{fig:learning_curve}
\end{figure}

The focus of our obstacle avoidance model is on recognizing and avoiding obstacles, instead of reasoning about the topology of the environment. Therefore, we train our agent in one empty rectangular room instead of HouseExpo dataset. At the beginning of each episode, our simulator randomly generates $n$ objects inside the room where $n\sim U(1,5)$ and each episode lasts $200$ steps. The laser range $l_l$ is $5$ m and the state length $l$ is $3$ m. The neuron network has $3$ convolutional layers as the configuration in \cite{Mnih2015Human}, followed by a Long-Short Term Memory (LSTM) with $128$ units. Proximal Policy Optimization (PPO) \cite{schulman2017proximal} is employed to train the network and the learning curve is depicted in Fig. \ref{fig:learning_curve}. 

\begin{table}[]
     \centering
     \caption{Action Commands Used in Simulation And Reality. $\Delta x$, $\Delta \theta$, $v$, $\omega$ Represent Linear Step Length (meter), Angular Step Length (degree), Linear Velocity (meter/sec) and Angular Velocity (degree/sec).}
        \begin{tabular}{c|ccc}
            \hline \hline 
              & forward & left rotation & right rotation \\
            \hline
            train ($\Delta x$, $\Delta \theta$) & (0.3, 0) & (0,10) & (0, -10)\\
            test ($v$, $\omega$)  & (0.2, 0) & (0, 40) & (0, -40)\\
            \hline
        \end{tabular}
     \label{tab:action}
 \end{table}

\begin{figure}
	\centering
	\includegraphics[height=0.25\textwidth]{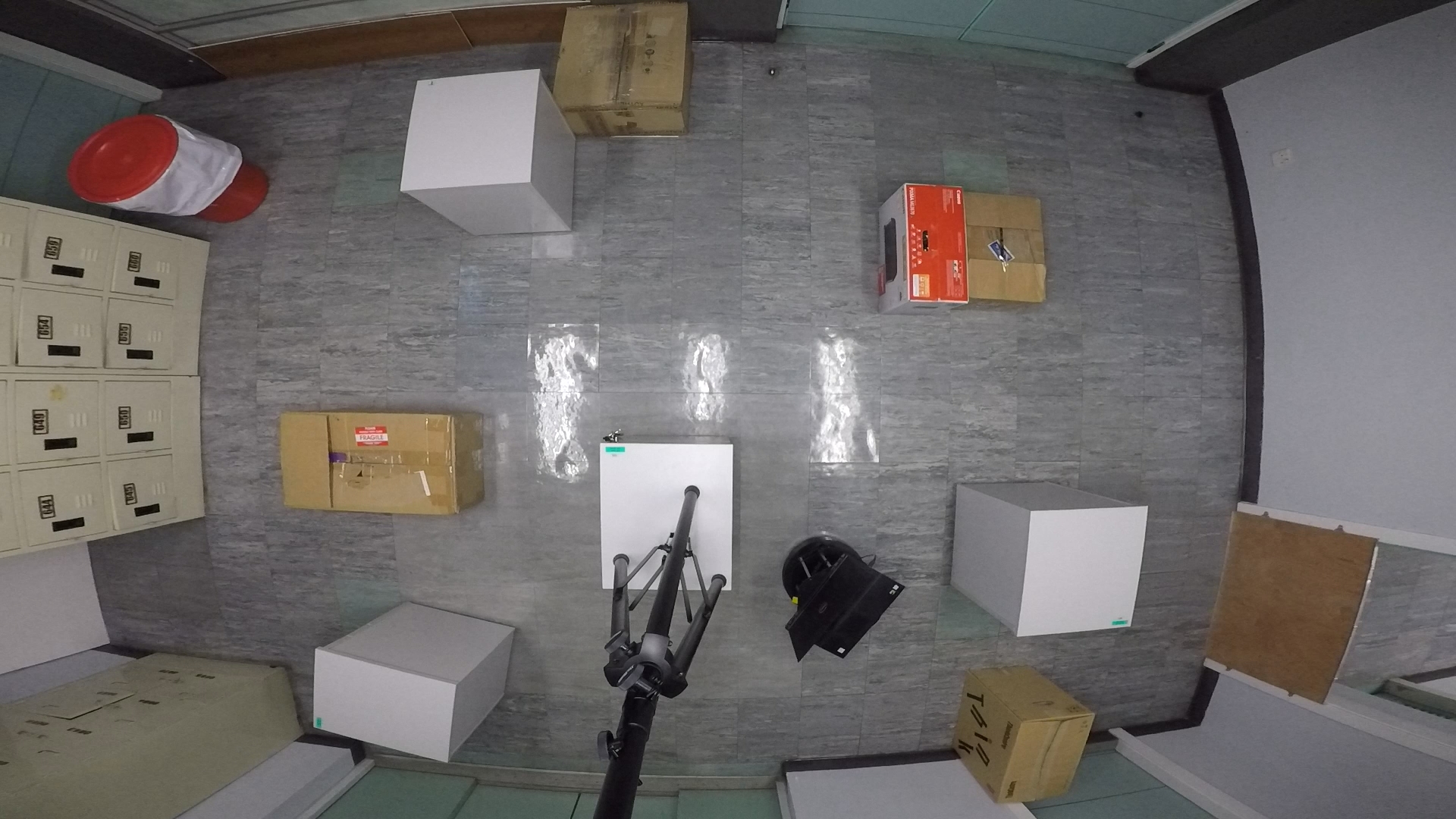}
	\caption{One test scene example. The room is $3.2\times5.5\text{m}^2$ and filled with objects.} 
	\label{fig:scene}
\end{figure}
\begin{table}[]
     \centering
     \caption{Obstacle Avoidance Performance in Real-World Experiments. $\#$Collisions is the total collisions happened in one scene over 10 episodes.}
        \begin{tabular}{c|ccccc}
            \hline \hline 
              & Scene 1 & Scene 2 & Scene 3 & Scene 4 & Scene 5  \\
            \hline
            $\#$Objects & 6 & 6 & 7 & 9 & 9  \\
            $\#$Collisions & 2 & 0 & 2 & 0 & 2 \\
            Mean(Traj. Len.)  & 18.41 & 17.98 & 18.08  & 15.62  & 14.67\\
            Var(Traj. Len.) & 0.70 & 4.31 & 2.01 & 21.78 & 15.25 \\
            \hline
        \end{tabular}
     \label{tab:durtion_dist}
 \end{table}

\begin{figure*}
	\centering
	\includegraphics[height=0.23\textwidth]{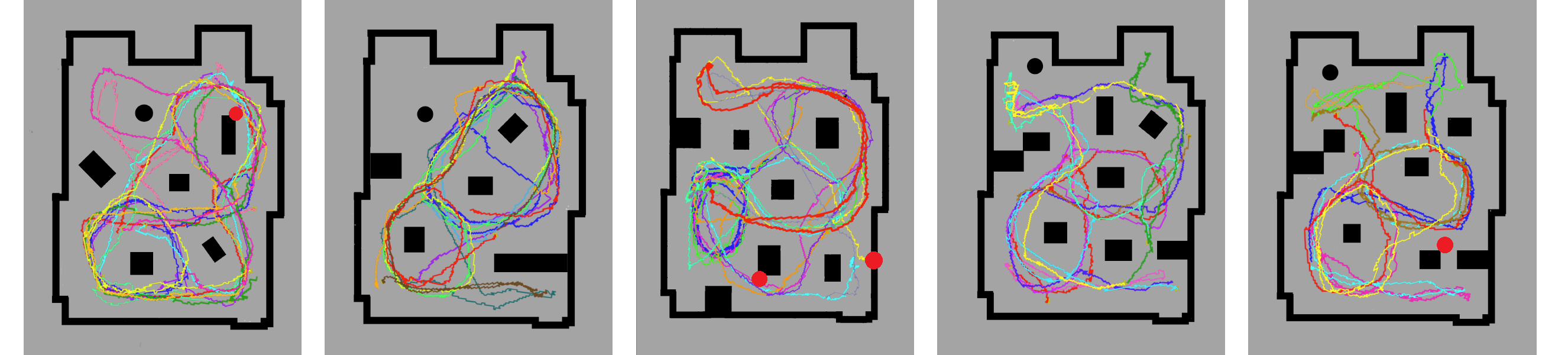}
	\caption{Trajectory demonstrations from Scene 1 to Scene 5 in real-world experiments. Each scene is tested for 10 times and trajectories are indicated with different colors. The collisions are indicated in red circles. The TurtleBot can traverse most of the free space while avoiding obstacles. } 
	\label{fig:trajectories}
\end{figure*}

Then we deploy the learned policy to a \emph{TurtleBot}. The gMapping algorithm is employed to construct the grid map, with the configuration consistent with that in the simulator including observation size, laser range, map resolution, etc. Furthermore, the actions in PseudoSLAM are discrete, so we map the action commands to continuous space as displayed in TABLE \ref{tab:action}. The codes run on a laptop with Intel Core i-5-8300H and Nvidia Geforce 1050Ti, achieving a $10$Hz control rate. We evaluate our trained model in a room with a size of $3.2$ $\times5.5\text{m}^2$ filled with objects. $5$ different object layouts are tested and in each scene, $10$ episodes with random starting points are evaluated. One example scene is shown in Fig.~\ref{fig:scene}. The time limit of each episode is $2$ minutes. In total, we conduct 50 experiments for 100 minutes. The experiments can be seen in the supplementary video.

We quantitatively evaluate our model in real-world scenes from two perspectives: the number of collisions and distance (trajectory length). The number of collisions is the summation of the times that the robot hit walls or obstacles in one scene over the 10 episodes, reflecting the basic ability to avoid obstacles. On the other hand, there are cases where the robot always rotates in place or just goes through a small region, which is unlikely to hit anything but not consistent with the goal. Thus we measure the trajectory length that the robot traverse. The experimental results are shown in TABLE \ref{tab:durtion_dist}. As we can see, the performance of the model is stable and only $2$ collisions happen in $3$ scenes. Another observation is the mean of the trajectory length decreases with the number of objects, reflecting that the robot is more careful in complicated environments and takes more action in adjusting its pose. 

Fig. \ref{fig:trajectories} gives the trajectories of the $10$ episodes in all $5$ test scenes. As we can see, our policy is robust and can be generalized to real-world scenarios. The robot can traverse most of the areas while keeping a safe distance to the objects. The robot can even reach some complex regions and move out of them, for example, the dead end at the bottom right corner in Scene $2$. This experiment demonstrates that the experience generated in PseudoSLAM can be applied to real-world situations and make the training process more efficient.
 
 \subsection{Autonomous Exploration}
 In this part, we demonstrate the effectiveness of topological information through robot exploration task. As justified in \cite{aydemir2012}, the spatial knowledge can be utilized to reason about unknown spaces in indoor environments and we use such knowledge to guide the exploration process. 
 
\begin{figure}
     \centering
	 \includegraphics[height=0.2\textwidth]{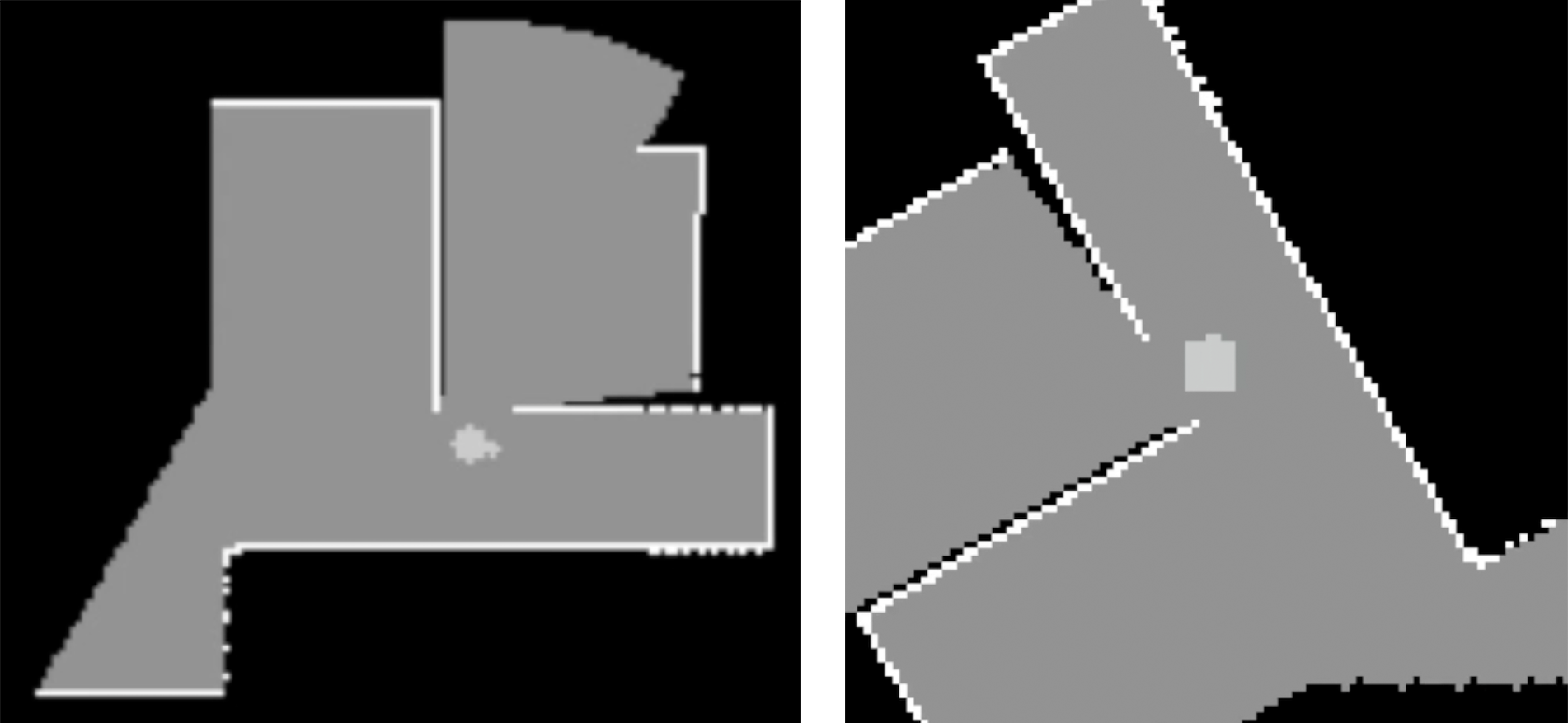}
     \caption{Global map and local observation example. Left figure shows a global map where an agent is exploring a house. Right figure is the input to the network. Uncertain areas, walls and free spaces are indicated in black, white and gray color, respectively. }
     \label{fig:global_local}
 \end{figure}

 Autonomous exploration refers to the process of searching for unknown areas. In our experiment, the robot is expected to discover as much area as possible to collect more information within a fixed time limit. Thus the reward function is defined as $r_t=\alpha r_s$, where $r_s$ is the newly discovered area at time $t$, encouraging the robot to move towards the unknown areas. Since the main focus of this experiment is on utilizing 2D layout information, all the training and testing houses are empty without adding any obstacles. 
 We split HouseExpo into training and testing set with $24,588$ and $10,538$ maps, respectively.
 The agent only observes a local rectangular map around it, as demonstrated in Fig.~\ref{fig:global_local}. Such an observation is rotation-invariant where the agent is always facing forward. We compare our method with a random policy where each action has an equal possibility of $\frac{1}{3}$ to be selected. 
 The network structure is the same as Section \ref{sec:OA} but the state length $l$ is $4$ m. We record the explored area within $200$ steps as the measurement of exploration.
 
 \begin{figure}
 	\centering
 	\includegraphics[width=0.5\textwidth]{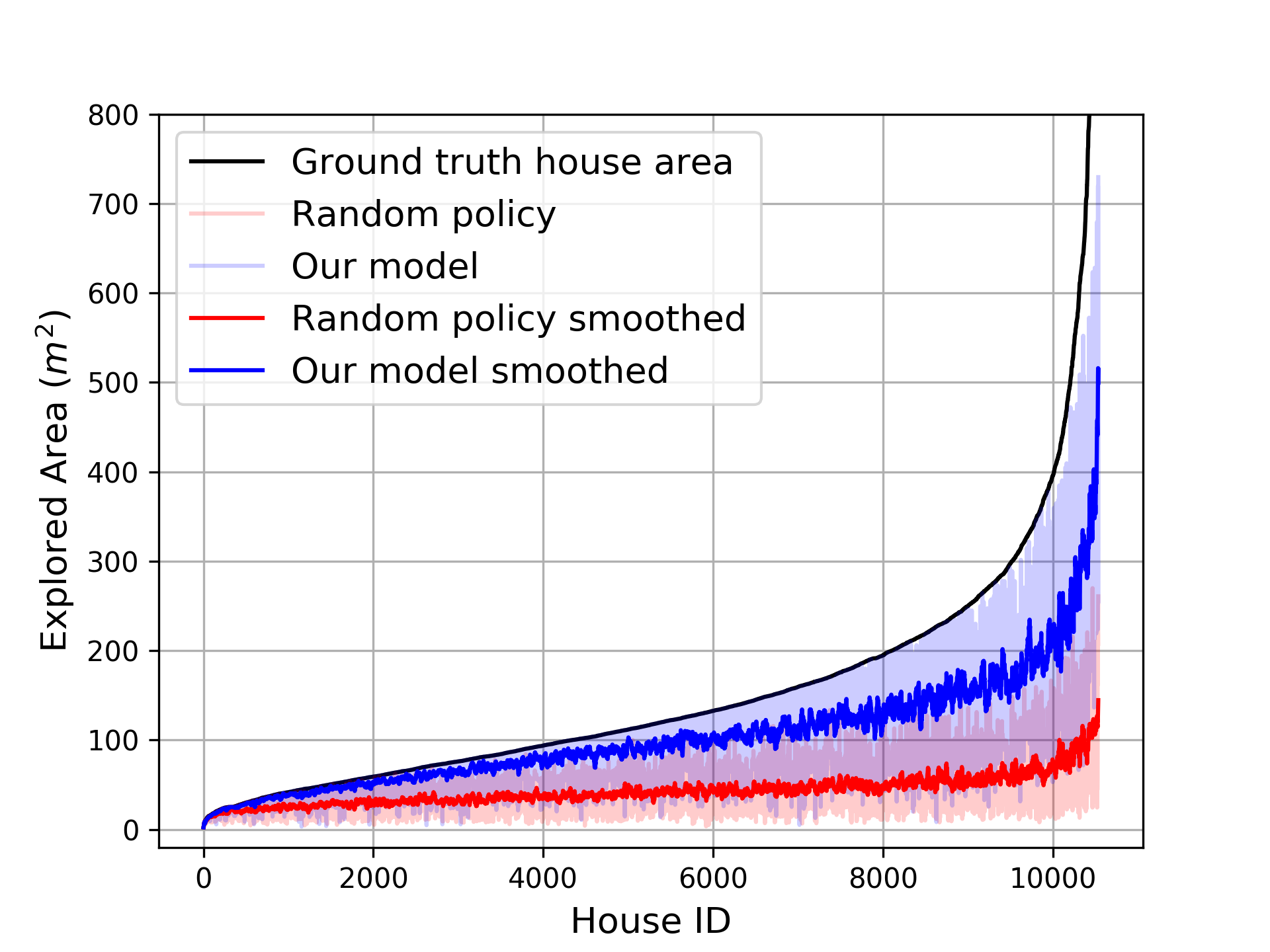}
 	\caption{Explored area in $10,538$ testing houses. The data has been sorted in ascending order of house areas.} \label{fig:ae}
 \end{figure}
 
 The simulation result is shown in Fig. \ref{fig:ae}. The solid curves represent smoothed results with a window size $20$. The House ID (x-axis) is sorted in ascending order according to their total areas. The groundtruth house area is indicated in black. By comparing our model with random policy, it is evident that our model performs much better with much higher traversed areas in all the $10,538$ houses, indicating that the spatial structures can be learned and transferred to new environments. 
  
 \section{CONCLUSIONS}
     In this paper, we build HouseExpo, a large-scale indoor layout dataset, and PseudoSLAM, an efficient simulation platform to facilitate applying learning-based methods to mobile robots. The effectiveness of our dataset and simulation platform is verified via simulation and real-world experiments.  
     
     Apart from tasks mentioned in Section \ref{sec:experiment}, we believe our dataset can also contribute to a number of other tasks, such as room segmentation, graph-bashed structure reasoning and path planning~\cite{9082624,9037111}, and scale up the diversity for algorithm evaluation. 
      
     At the same time, there is some future work to be done. One concern is how to optimally use the topological information. In our autonomous exploration experiment, only the local observation (a local map around the robot) is utilized. Since the sizes of houses vary a lot, it is impractical to directly feed the whole global map into convolutional neuron networks. It could be investigated how to represent the topology information. Another direction is incorporating further levels of control hierarchies. Take autonomous exploration problem as an example, the data-driven approach makes a long-term plan and designates a goal based on its experience and a local planner then plans a path and drives the robot to the goal. 
 
\section*{ACKNOWLEDGMENT}
This project is supported by the Hong Kong RGC GRF grants $\#$14200618 awarded to Max Q.-H. Meng.

\bibliographystyle{IEEEtran}
\bibliography{iros2018}

\end{document}